\UseRawInputEncoding
\documentclass[letterpaper, 10 pt, journal, twoside]{IEEEtran} 

\title{A Nonlinear MPC Framework for Loco-Manipulation of Quadrupedal Robots with Non-Negligible Manipulator Dynamics}

\usepackage{amsmath,amssymb,amsfonts}
\usepackage{cite}

\usepackage{graphicx} 
\usepackage{times} 
\usepackage{xcolor}
\usepackage{balance}
\usepackage{multicol}
\usepackage{array}
\usepackage{multirow}
\usepackage{booktabs}
\usepackage[colorlinks=true, allcolors=blue]{hyperref}

\newcommand{\Real}{\mathbb{R}}
\newcommand{\col}{\textrm{col}}
\newcommand{\Integer}{\mathbb{Z}_{\geq0}}
\newcommand{\diag}{\textrm{diag}}
\newcommand{\skews}{\mathbb S}
\newcommand{\identity}{\mathbb I}
\newcommand{\des}{\textrm{des}}

\newcommand{\SRB}{\textrm{SRB}}
\newcommand{\arm}{\textrm{arm}}
\newcommand{\inte}{\textrm{int}}
\newcommand{\net}{\textrm{net}}
\newcommand{\SO}{\textrm{SO}(3)}
\newcommand{\so}{\mathfrak{so}(3)}
\newcommand{\terminal}{\textrm{terminal}}
\newcommand{\stage}{\textrm{stage}}
\newcommand{\reff}{\textrm{ref}}
\newcommand{\full}{\textrm{full}}

\usepackage{etoolbox}

\AtBeginEnvironment{equation}{\vspace{-3pt}}
\AtEndEnvironment{equation}{\vspace{-3pt}}
\AtBeginEnvironment{align}{\vspace{-3pt}}
\AtEndEnvironment{align}{\vspace{-3pt}}
\AtBeginEnvironment{alignat}{\vspace{-3pt}}
\AtEndEnvironment{alignat}{\vspace{-3pt}}

\newcounter{definitionCounter}
\newcounter{theoremCounter}
\newcounter{lemmaCounter}

\author{Ruturaj Sambhus$^{1*}$, 
Kapi Ketan Mehta$^{1*}$, Ali MirMohammad Sadeghi$^{1*}$, 
Basit Muhammad Imran$^{1}$, Jeeseop Kim$^{2}$, Taizoon Chunawala$^{1}$, Vittorio Pastore$^{1}$, Sujith Vijayan$^{3}$, and Kaveh Akbari Hamed$^{1}$

\thanks{$^{*}$These authors contributed equally to this work.}
\thanks{The work of R.~Sambhus, S.~Vijayan, and K.~Akbari Hamed is partially supported by the National Science Foundation (NSF) under Grant 2423725.}
\thanks{$^{1}$R.~Sambhus, K.~K.~Mehta, A.~M.~Sadeghi, B.~Imran, T.~Chunawala,  V.~Pastore, and K.~Akbari Hamed (\textit{Corresponding Author}) are with the Department of Mechanical Engineering, Virginia Tech, Blacksburg, VA 24061, USA, {\tt\small \{ruturajsambhus, kmehta4, alimmsadeghi, basit, taizoonc, vittoriopastore, kavehakbarihamed\}@vt.edu}}
\thanks{$^{2}$J.~Kim is with California Institute of Technology, Pasadena, CA 91125, USA, {\tt\small jeeseop@caltech.edu}}
\thanks{$^{3}$S.~Vijayan is with the School of Neuroscience, Virginia Tech, Blacksburg, VA 24061, USA, {\tt\small neuron99@vt.edu}}
}

\begin{document}
\maketitle


\begin{abstract}
Model predictive control (MPC) combined with reduced-order template models has emerged as a powerful tool for trajectory optimization in dynamic legged locomotion. However, loco-manipulation tasks performed by legged robots introduce additional complexity, necessitating computationally efficient MPC algorithms capable of handling high-degree-of-freedom (DoF) models. This letter presents a computationally efficient nonlinear MPC (NMPC) framework tailored for loco-manipulation tasks of quadrupedal robots equipped with robotic manipulators whose dynamics are non-negligible relative to those of the quadruped. The proposed framework adopts a decomposition strategy that couples locomotion template models---such as the single rigid body (SRB) model---with a full-order dynamic model of the robotic manipulator for torque-level control. This decomposition enables efficient real-time solution of the NMPC problem in a receding horizon fashion at 60 Hz. The optimal state and input trajectories generated by the NMPC for locomotion are tracked by a low-level nonlinear whole-body controller (WBC) running at 500 Hz, while the optimal torque commands for the manipulator are directly applied. The layered control architecture is validated through extensive numerical simulations and hardware experiments on a 15-kg Unitree Go2 quadrupedal robot augmented with a 4.4-kg 4-DoF Kinova arm. Given that the Kinova arm dynamics are non-negligible relative to the Go2 base, the proposed NMPC framework demonstrates robust stability in performing diverse loco-manipulation tasks, effectively handling external disturbances, payload variations, and uneven terrain.
\end{abstract}



\vspace{-1em}
\section{Introduction}
\label{sec:Intro}

Model predictive control (MPC), when combined with reduced-order template models, has emerged as a powerful framework for trajectory optimization in dynamic legged locomotion. Reduced-order models \cite{Full_Koditschek_Template} provide low-dimensional abstractions of complex, nonlinear locomotion systems. Common examples include the linear inverted pendulum (LIP) model \cite{kajita19991LIP} and its extensions, such as the angular momentum LIP \cite{ALIP}, spring-loaded inverted pendulum (SLIP) \cite{SLIP}, vertical SLIP \cite{vLIP_Sreenath}, and hybrid LIP \cite{HLIP_Ames}, as well as other templates like centroidal dynamics \cite{orin2013centroidal} and the single rigid body (SRB) model \cite{Kim_Wensing_Convex_MPC_01,Wensing_VBL_HJB,Abhishek_Hae-Won_TRO,Leila_Hamed_RAL,pandala2022robust}. However, loco-manipulation tasks---performed by legged robots equipped with articulated manipulators---introduce new layers of complexity. These tasks require MPC algorithms that are not only computationally efficient but also capable of coordinating high-degree-of-freedom (DoF) motions under dynamic constraints.

\begin{figure}[t]
    \centering
    \includegraphics[width=\linewidth]{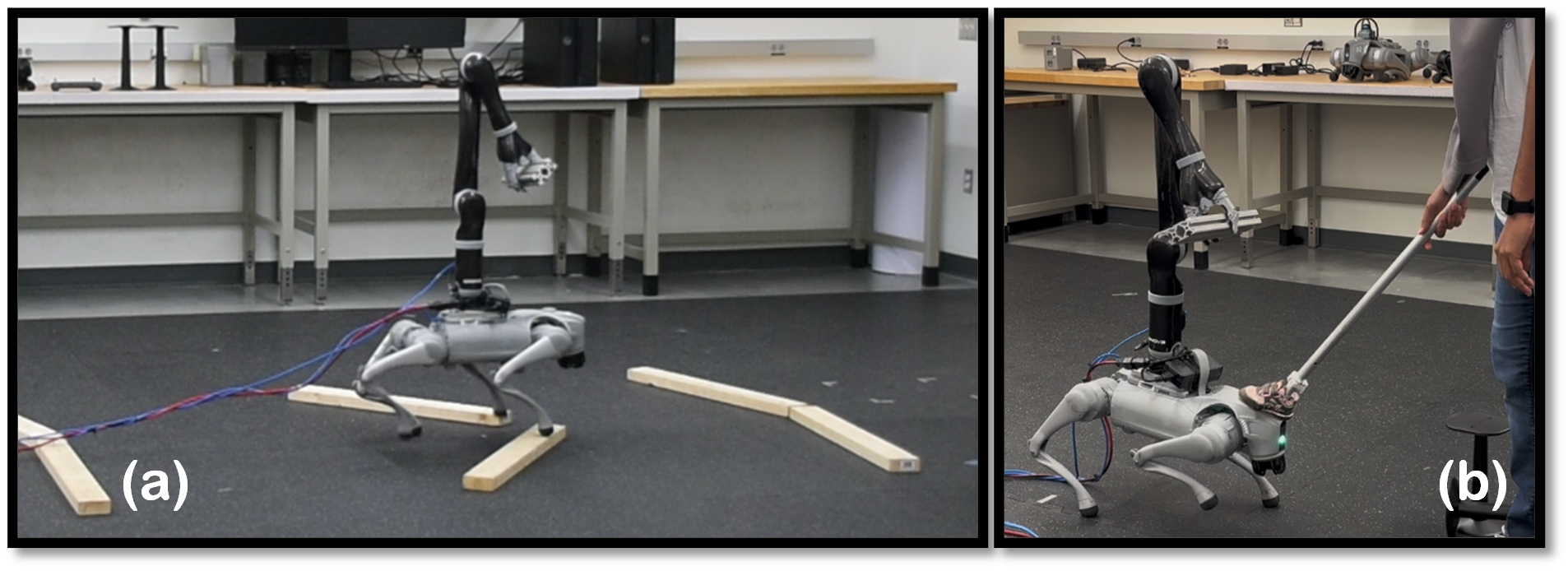}
    \vspace{-2em}
    \caption{Robust loco-manipulation by the Unitree Go2 quadrupedal robot augmented with a 4-DoF Kinova arm, (a) trotting over wooden blocks while carrying a 1 kg payload and (b) withstanding an external push disturbance.}
    \vspace{-1.5em}
    \label{fig:Onesnapshot}
\end{figure}

Loco-manipulation planning can be formulated as a multi-contact trajectory optimization and control problem. Recent theoretical and technological advances have enabled legged platforms---especially quadrupeds---to perform increasingly complex tasks such as throwing objects \cite{Murphy2012HighDD}, opening spring-loaded doors \cite{Sleiman_RAL,Sleiman_Science}, turning hand wheels \cite{RoloMa_Ferrolho,Sleiman_Science}, and pushing or pulling heavy payloads \cite{RoloMa_Ferrolho,Sleiman_RAL,ALMA_Bellicoso,MA_RAL}. While whole-body planning algorithms have been proposed to support dynamic loco-manipulation by leveraging centroidal dynamics and full-body kinematics \cite{Sleiman_RAL}, many such approaches approximate or neglect the full-order dynamics of the manipulator itself \cite{Dai_WBC_Kinematics}. This simplification becomes limiting in cases where arm dynamics are non-negligible, such as when integrating a 4.4-kg, 4-DoF Kinova arm with a 15-kg Unitree Go2 quadruped, as illustrated in Fig.~\ref{fig:Onesnapshot}.

The \textit{overarching goal} of this paper is to present a unified and computationally efficient nonlinear MPC (NMPC) framework that couples locomotion template models with full-order manipulator dynamics via a model decomposition strategy. The proposed framework enables robust and torque-controlled loco-manipulation on rough terrains and under disturbances, particularly when the manipulator dynamics are significant relative to those of the base quadruped.


\begin{figure*}[t]
    \centering
    \includegraphics[width=\linewidth]{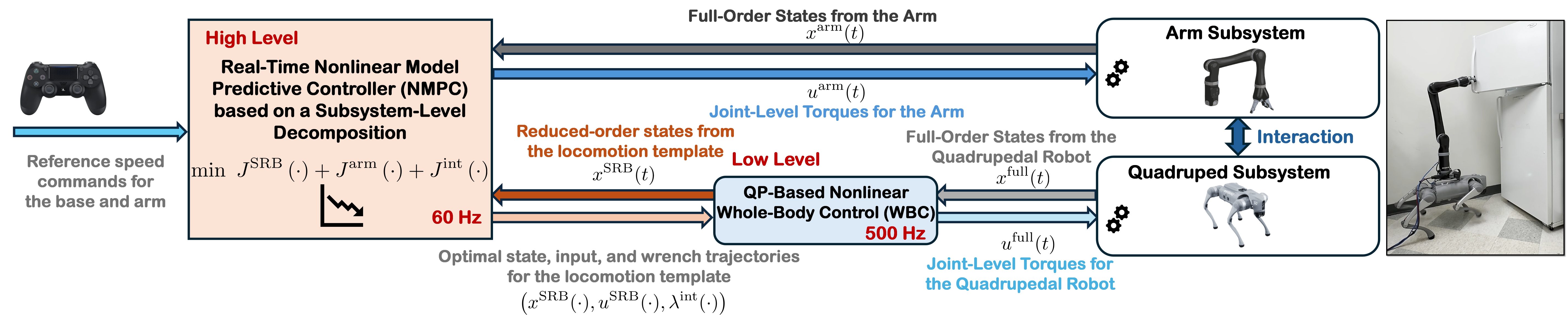}
    \vspace{-2.3em}
    \caption{Overview of the proposed layered control framework, which consists of a high-level NMPC trajectory planner---based on a decomposition approach---for both the locomotion template model and the full-order arm model, along with a low-level nonlinear WBC for the full-order locomotion model.}
    \vspace{-1.4em}
    \label{fig:Overview}
\end{figure*}


\vspace{-1em}
\subsection{Related Work}
\label{sec:Related_work}

A unified MPC framework for whole-body loco-manipulation was introduced in \cite{Sleiman_RAL}, where real-time MPC is applied to the robot's centroidal dynamics and full-body kinematics (i.e., the kino-centroidal model) while explicitly accounting for object dynamics. This approach was demonstrated on ALMA, a robotic system comprising the 50-kg ANYmal quadruped equipped with an 8-kg, 4-DoF DynaArm. Building on this work, \cite{Sleiman_Science} proposes a sampling-based bilevel optimization framework to compute optimal multi-modal action sequences for goal-reaching tasks. A hierarchical two-stage planner for ALMA is introduced in \cite{Mittal_artculated}, which integrates object-centric and agent-centric planning to coordinate loco-manipulation behaviors. To enhance robustness under external disturbances, \cite{RoloMa_Ferrolho} presents an offline nonlinear program (NLP) for generating dynamic motions, using ANYmal with a 6-DoF Kinova arm. A distributed whole-body MPC is developed in \cite{Semini_ADMM} and validated through simulations for quadrupedal locomotion with concurrent arm motion.


Learning-based methods have also shown promise in this domain. In \cite{MA_RAL}, a reinforcement learning (RL) locomotion policy is combined with a model-based MPC controller for manipulation, achieving robust performance on rough terrains. A unified RL-based policy for whole-body visuomotor control in badminton-playing tasks is presented in \cite{Ma_Badminton}. In \cite{fu_deepWBC}, an RL-based control framework is developed for loco-manipulation using a Unitree Go1 robot equipped with a lightweight 6-DoF WidowX 250 S arm.  To improve resilience, \cite{Ma_Arm_Assisted_Reduction} develops an RL-based controller that enables fall damage mitigation and recovery in legged robots equipped with manipulators. In \cite{Zimmermann_GoFetch}, a numerical model for dynamic grasping is developed for the Spova platform---a Spot robot paired with a 7-DoF Kinova arm---where model parameters are fitted to experimental data and used for trajectory optimization. An adaptive learning-based framework for long-horizon skill coordination in pick-and-place tasks with Spot is presented in \cite{yokoyama2023asc}. Additionally, \cite{pan2025roboduet} proposes a cooperative RL-based whole-body control policy for loco-manipulation using a Unitree Go1 with a lightweight ARX5 arm, and \cite{zhang2024learning} develops a learning-based controller for door opening and traversal tasks. 

Despite these advancements, most MPC-based approaches remain centered around kino-centroidal models, rather than incorporating full-order manipulator dynamics. This simplification can compromise robustness and stability in loco-manipulation tasks involving quadrupeds with relatively heavy arms---for example, the Unitree Go2 integrated with a 4-DoF Kinova arm, where the arm-to-base mass and height ratios reach 29\% and 216\%, respectively. This paper seeks to address the following \textit{fundamental question}: How can full-order manipulator dynamics be systematically integrated with locomotion template models to enable real-time, efficient, and robust NMPC algorithms for loco-manipulation?


\vspace{-1em}
\subsection{Contributions}
\label{sec:Contributions}

To address the above question, the paper aims to develop a unified and computationally efficient NMPC framework for loco-manipulation of quadrupedal robots, where the manipulator dynamics are non-negligible relative to the simplified template models typically used for locomotion. The \textit{contributions} of this work are as follows. We introduce a \textit{decomposition strategy} that couples locomotion template models---such as the SRB model---with full-order dynamic models of manipulators via rigid holonomic constraints. This approach preserves the effectiveness of locomotion templates like SRB for dynamic quadrupedal locomotion while systematically integrating arm dynamics for manipulation tasks. The decomposition enables a unified, real-time NMPC formulation that optimizes both the reduced-order states and inputs of the locomotion template model and the full-order states and joint torques of the arm, operating at 60 Hz. It also facilitates subsystem-specific decomposition of NMPC into locomotion and manipulation components while explicitly regulating the interaction wrench between them. The optimal state and input trajectories for locomotion are tracked by a low-level nonlinear whole-body controller (WBC), based on quadratic programming (QP) and virtual constraints \cite{Jessy_Book, Randy_Paper_LCSS}, running at 500 Hz, while the optimal torque commands for the manipulator are applied directly (see Fig. \ref{fig:Overview}). 

The effectiveness of the proposed layered optimal control architecture is validated through extensive simulations and hardware experiments on a 15 kg Unitree Go2 quadrupedal robot equipped with a 4.4 kg, 4-DoF torque-controlled Kinova arm, selected for its mass and torque control characteristics (see Fig. \ref{fig:Onesnapshot}). Given that the arm dynamics are non-negligible relative to the Go2 base, the proposed NMPC framework demonstrates robust stability in executing diverse loco-manipulation tasks under external disturbances, payload variations, uneven terrain, and tracking time-varying reference trajectories. These tasks include locomotion over rough terrain while manipulating unknown objects and withstanding external push disturbances; pushing and pulling unmodeled objects such as a wagon cart, with up to 125\% mass uncertainty relative to the robot's base mass; opening doors, drawers, and fridge doors (see Fig. \ref{fig:Overview}); and performing pick-and-place operations. To further evaluate the efficacy of the proposed decomposition-based NMPC approach, we conduct numerical simulations on 200 randomly generated terrains and compare the results against the baseline kino-centroidal NMPC approach from \cite{Sleiman_RAL}. Our proposed method achieves a 58\% higher success rate of robust locomotion than the baseline approach. 

Prior works on loco-manipulation with small quadrupeds like the Go1, including RL-based approaches \cite{fu_deepWBC,pan2025roboduet}, focus on lightweight, position-controlled arms. The adaptive MPC in \cite{Leila_Hamed_RAL} tackled static payload transport by A1 robots on rough terrain with up to 91\% model uncertainty, but under low-dynamic conditions. However, the Go2's nominal 8-kg static payload capacity does not account for dynamic, tall manipulators like the Kinova arm. To the best of our knowledge, no model-based control method has addressed dynamic loco-manipulation with full-order, torque-controllable, heavier arms on small quadrupeds. This paper introduces the first such algorithm, enabling dynamic loco-manipulation near the physical limits of the platform. 


\vspace{-1em}
\section{Dynamic Model for Loco-Manipulation}
\label{sec:Template_Model}

The objective of this section is to derive a dynamic model that unifies locomotion template models with full-order manipulator dynamics through a decomposition-based approach. This integrated model forms the foundation for the computationally efficient NMPC framework for loco-manipulation, introduced in Section~\ref{sec:NMPC}.

\textbf{Template for Locomotion:} We adopt the simplified SRB dynamics of the robot as the template model for locomotion, with the state vector defined as $x^{\SRB}:=\col(p,\dot{p},\theta,\omega)\in\Real^{12}$, where $\col(\cdot)$ denotes the column operator. Here, $p\in\Real^{3}$ denotes the Cartesian coordinates of the robot's center of mass (CoM), $\theta\in\Real^{3}$ represents the Euler angles (roll, pitch, and yaw) describing the body orientation with respect to the world frame, and $\omega\in\Real^{3}$ is the angular velocity of the body, also expressed in the world frame. The corresponding rotation matrix is denoted by $R(\theta)\in\SO$. The SRB dynamics are governed by the following equations of motion:
\begin{equation}\label{eq:SRB_dyn}
    \Sigma^{\SRB}: \begin{cases}
    \ddot{p}      = \frac{f^{\net}}{m} - g_0 \\
    \dot{\theta}  = A(\theta)\,\omega \\
    \dot{\omega}  = I^{-1} \left(\tau^{\net} - \skews(\omega) \, I\, \omega \right),
\end{cases}
\end{equation}
where $m$ denotes the mass of the quadrupedal robot, $g_{0}\in\Real^{3}$ represents the gravitational vector, $I\in\Real^{3\times3}$ is the inertia expressed in the world frame as $I=R(\theta)\,I^{B}\,R^\top(\theta)$, with $I^{B}$ being the inertia in the body frame, and $\skews(\cdot):\Real^3 \rightarrow \so$ denotes the skew-symmetric matrix operator with the property $\skews(a)\,b=a\times b$ for every $a,b\in\Real^{3}$. In our formulation, the transformation matrix $A(\theta)\in\Real^{3\times3}$ maps the angular velocity in the world frame to the time derivative of Euler angles.

The net force and torque acting on the CoM are computed as:
\begin{equation}\label{eq:net_wrench}
    \begin{bmatrix}
    f^{\net}\\
    \tau^{\net}
    \end{bmatrix}:=\sum_{\ell\in\mathcal{C}} \begin{bmatrix}
        f^{\ell}\\
        \skews(r^{\ell})\,f^{\ell}
    \end{bmatrix} + \begin{bmatrix}
        f^{\inte}\\
        \tau^{\inte}
    \end{bmatrix} + \begin{bmatrix}
        0\\
        \skews(r^{\inte})\,f^{\inte}
    \end{bmatrix},
\end{equation}
where $\ell\in\mathcal{C}$ indexes the stance feet, $\mathcal{C}$ denotes the set of stance feet, $f^{\ell}\in\Real^{3}$ is the ground reaction force (GRF) at stance foot $\ell$, and $r^{\ell}\in\Real^{3}$ is the vector from foot $\ell$ to the CoM. The term  $\lambda^{\inte}:=\col(f^{\inte},\tau^{\inte})\in\Real^{6}$ represents the interaction wrench between the SRB and the arm dynamics. The induced torque due to the interaction force $f^{\inte}$ is captured by the term $\skews(r^{\inte})\,f^{\inte}$, where $r^{\inte}\in\Real^{3}$ denotes the vector from the CoM to the point of interaction. 

By treating the GRFs as control inputs for the SRB dynamics---denoted by $u^{\SRB}$---the equations of motion in \eqref{eq:SRB_dyn} and \eqref{eq:net_wrench} can be discretized using Euler's method and expressed in a nonlinear state-space form as:
\begin{equation}\label{eq:SRB_state_space}
    x^{\SRB}(t+1)=f^{\SRB}\left(x^{\SRB}(t),u^{\SRB}(t),\lambda^{\inte}(t)\right), \,\, t\in\Integer,
\end{equation}
where $t$ denotes the discrete time and $\Integer:=\{0,1,\cdots\}$. We remark that model \eqref{eq:SRB_state_space} is valid if $u^{\SRB}(t)\in\mathcal{U}^{\SRB}$, where $\mathcal{U}^{\SRB}$ represents the friction cone. 

\textbf{Manipulation Model:} Since our objective is to perform torque-level control of the arm within the proposed NMPC framework, we incorporate a floating-base dynamic model of the arm into the optimal control formulation. A floating-base coordinate system is adopted for the arm to enable computation of the interaction wrench $\lambda^{\inte}$. Let $q^{\arm}:=\col(q^{b},q^{s})\in\Real^{10}$ denote the floating-base coordinates of the 4-DoF Kinova arm, where $q^{b}\in\Real^{6}$ corresponds to the base coordinates, and  $q^{s}\in\Real^{4}$ represents the shape variables. Specifically, $q^{b}:=\col(p^{b},\theta^{b})$, where $p^{b}\in\Real^{3}$ denotes the Cartesian coordinates of the base of the arm, and $\theta^{b}$ represents the corresponding Euler angles. 

The arm's equations of motion are derived using the Euler-Lagrange formalism in conjunction with the principle of virtual work and Newton's third law, resulting in:
\begin{equation}\label{eq:arm_dyn}
    D\left(q^{\arm}\right) + H\left(q^{\arm},\dot{q}^{\arm}\right) = B\,u^{\arm} - J^{\top}\left(q^{\arm}\right)\,\lambda^{\inte},
\end{equation}
where $D(q^{\arm})\in\Real^{10\times10}$ is the positive-definite mass-inertia matrix, $H(q^{\arm},\dot{q}^{\arm})\in\Real^{10}$ includes Coriolis, centrifugal, and gravitational effects, $B\in\Real^{10\times4}$ is the input distribution matrix, $u^{\arm}\in\mathcal{U}^{\arm}\subset\Real^{4}$ is the vector of joint-level torques, where $\mathcal{U}^{\arm}$ denotes the set of admissible torques. The Jacobian matrix  $J(q^{\arm})\in\Real^{6\times10}$ maps the interaction wrench into the dynamics. The equations of motion in \eqref{eq:arm_dyn} can be discretized and rewritten in a nonlinear state-space form as:
\begin{equation}\label{eq:arm_state_space}
    x^{\arm}(t+1)=f^{\arm}\left(x^{\arm}(t),u^{\arm}(t),\lambda^{\inte}(t)\right),\,\, t\in\Integer,
\end{equation}
where $x^{\arm}:=\col(q^{\arm},\dot{q}^{\arm})\in\Real^{20}$ denotes the state vector of the arm subsystem.

\textbf{Holonomic Constraints:} In addition to the nonlinear state equations of the SRB and arm subsystems given in \eqref{eq:SRB_state_space} and \eqref{eq:arm_state_space}, we incorporate the rigid contact condition at the interaction point as a holonomic constraint. This constraint enforces that the base of the arm remains in rigid contact with the SRB, thereby enabling the computation of the interaction wrench (i.e, Lagrange multipliers) $\lambda^{\inte}$. Specifically, at the interaction point, we impose the following constraints:
\begin{equation}\label{eq:holonomic_const}
    p+R(\theta)\,d = p^{b} \quad \textrm{and} \quad \theta=\theta^{b},
\end{equation}
where $d\in\Real^{3}$ is the fixed offset vector from the CoM of the SRB to the interaction point. The right-hand sides, $p^{b}$ and $\theta^{b}$, denote the position and orientation of the base of the arm with respect to the world frame, expressed in the arm's floating-base coordinates. The left-hand sides express the same quantities derived from the SRB state using the SRB frame. The holonomic constraints in \eqref{eq:holonomic_const} can be expressed in a compact form as follows: 
\begin{equation}
    \varphi(p,\theta,q^{\arm}) = 0.
\end{equation}

To enforce these constraints during motion, we differentiate  \eqref{eq:holonomic_const} twice with respect to time. Using the relation $\dot{\theta}=A(\theta)\,\omega$ from the SRB dynamics \eqref{eq:SRB_dyn}, the first-order derivatives are:
\begin{alignat}{4}
\dot{p}+\frac{\partial }{\partial \theta} \left(R\,d\right)\,A\,\omega = \dot{p}^{b}\\
A\,\omega = \dot{\theta}^{b}.
\end{alignat}
Differentiating again yields the second-order constraint equations:
\begin{alignat}{4}
\ddot{p} + \frac{\partial}{\partial \theta} \left(\frac{\partial}{\partial \theta}\left(R\,d\right)A\,\omega\right)A\,\omega + \frac{\partial }{\partial \theta} \left(R\,d\right)\,A\,\dot{\omega} = \ddot{p}^{b}\\
A\,\dot{\omega} + \frac{\partial}{\partial \theta} \left(A\,\omega\right)A\,\omega = \ddot{\theta}^{b}.
\end{alignat}
Substituting $\ddot{p}$, $\dot{\omega}$, $\ddot{p}^{b}$ and $\ddot{\theta}^{b}$ from the continuous-time SRB and arm dynamics in \eqref{eq:SRB_dyn} and \eqref{eq:arm_dyn} yields a dynamic consistency condition, which can be expressed as a function of the control inputs and the interaction wrench:
\begin{equation}\label{eq:holonomic_const_diff}
    \ddot{\varphi}\left(x^{\SRB}(t),x^{\arm}(t),u^{\SRB}(t),u^{\arm}(t),\lambda^{\inte}(t)\right)=0
\end{equation}
for every $t\in\Integer$. 

\textbf{Computation of the Interaction Wrench:} From \eqref{eq:holonomic_const_diff}, one may solve for the interaction wrench $\lambda^{\inte}$ and substitute the result into the state equations of the SRB and arm subsystems in \eqref{eq:SRB_state_space} and \eqref{eq:arm_state_space}. However, this approach introduces additional nonlinearity into the system dynamics, thereby increasing the complexity of the resulting optimal control problem.  Alternatively, we adopt a computationally effective strategy by incorporating the state equations \eqref{eq:SRB_state_space} and \eqref{eq:arm_state_space} as constraints in the optimal control formulation, together with the holonomic constraint \eqref{eq:holonomic_const_diff}. In this formulation, the decision variables of the nonlinear MPC problem include the trajectories of the states, control inputs, and the interaction wrench over the control horizon---denoted by $x^{\SRB}(\cdot)$, $x^{\arm}(\cdot)$, $u^{\SRB}(\cdot)$, $u^{\arm}(\cdot)$, and $\lambda^{\inte}(\cdot)$. This approach allows the NMPC to solve for the interaction wrench implicitly by enforcing the holonomic constraint \eqref{eq:holonomic_const_diff} as an equality constraint. Notably, the SRB and arm dynamics remain defined in terms of their own local states and control variables, and are coupled only through the holonomic constraint \eqref{eq:holonomic_const_diff}, which governs the interaction at the contact point.

\textbf{Problem Statement:} Our objective is to develop a computationally efficient and unified NMPC framework that simultaneously performs trajectory optimization for both the SRB and floating-base arm dynamics, as described in \eqref{eq:SRB_state_space} and \eqref{eq:arm_state_space}. The formulation is subject to the coupling equality constraint given in \eqref{eq:holonomic_const_diff}, ensuring coordinated and robustly stable loco-manipulation behavior.


\vspace{-1em}
\section{Layered Control Algorithm}
\label{sec:Layered_Control}

This section aims to present the proposed layered structure for trajectory optimization and control of loco-manipulation tasks. At the high level, a unified NMPC framework optimizes the state trajectories of both the SRB and the arm subsystems at 60 Hz. It simultaneously computes the optimal GRFs for locomotion and joint torques for manipulation. The resulting optimal state and GRF trajectories for the SRB are subsequently relayed to a low-level nonlinear WBC running at 500 Hz, which synthesizes the corresponding joint-level torques to realize the locomotion behavior (see Fig. \ref{fig:Overview}).


\vspace{-1em}
\subsection{Unified NMPC Formulation for Loco-Manipulation}
\label{sec:NMPC}

To address the problem in Section \ref{sec:Template_Model}, we propose the following real-time, unified NMPC formulation:
\begin{alignat}{4}
    &\min_{(\xi^{\SRB}(\cdot),\xi^{\arm}(\cdot),\lambda^{\inte}(\cdot))} \!\!\!\!\! &&J^{\SRB}\left(\xi^{\SRB}(\cdot)\right) + J^{\arm}\left(\xi^{\arm}(\cdot)\right) + J^{\inte}\left(\lambda^{\inte}(\cdot)\right)\nonumber\\
    & \quad \quad \quad \textrm{s.t.} && x^{\SRB}_{t+k+1|t} = f^{\SRB}\left(x^{\SRB}_{t+k|t},u^{\SRB}_{t+k|t},\lambda^{\inte}_{t+k|t}\right)\nonumber\\
    & && x^{\arm}_{t+k+1|t} = f^{\arm}\left(x^{\arm}_{t+k|t},u^{\arm}_{t+k|t},\lambda^{\inte}_{t+k|t}\right)\nonumber\\
    & && \ddot{\varphi}\left(x^{\SRB}_{t+k|t},x^{\arm}_{t+k|t},u^{\SRB}_{t+k|t},u^{\arm}_{t+k|t},\lambda^{\inte}_{t+k|t}\right)=0\nonumber\\
    & && E\,u^{\SRB}_{t+k|t}=0, \quad k=0,1,\cdots,N-1,\nonumber\\
    & && u^{\SRB}_{t+k|t}\in\mathcal{U}^{\SRB},\,\,\, u^{\arm}_{t+k|t}\in\mathcal{U}^{\arm}.
    \label{eq:unified_NMPC}
\end{alignat}
Here, $N$ represents the control horizon, and the optimization variables are defined as $(\xi^{\SRB}(\cdot),\xi^{\arm}(\cdot),\lambda^{\inte}(\cdot))$, where 
$\xi^{\SRB}(\cdot):=\col(x^{\SRB}(\cdot),u^{\SRB}(\cdot))\in\Real^{24N}$ and $\xi^{\arm}(\cdot):=\col(x^{\arm}(\cdot),u^{\arm}(\cdot))\in\Real^{24N}$ represent the state and control trajectories of the SRB and arm subsystems, respectively, over the prediction horizon. The variable $\lambda^{\inte}(\cdot)\in\Real^{6N}$ denotes the trajectory of the interaction wrenches. In our notation, the subscript $t+k|t$ denotes the predicted value at time $t+k$, computed at time $t$, for all $0\leq k \leq N-1$. The cost function is composed of subsystem-specific quadratic terms, $J^{\SRB}(\xi^{\SRB}(\cdot))$ and $J^{\arm}(\xi^{\arm}(\cdot))$, along with a penalty on the interaction wrench $J^{\inte}(\lambda^{\inte}(\cdot))$, which will be discussed in detail later. The equality constraints encode the discrete-time dynamics of the SRB and arm subsystems, along with the holonomic constraint $\ddot{\varphi}(\cdot)=0$, which enforces the rigid connection between them. Additionally, the available GRFs during locomotion are captured by the constraint $E\,u^{\SRB}_{t+k|t}=0$, where each entry of $E$ is set to $1$ if the corresponding component of $u^{\SRB}\in\Real^{12}$ is unavailable (e.g., due to swing phase) and $0$ otherwise. The inequality constraints encode the friction cone conditions for the SRB subsystem and ensure torque admissibility for the arm subsystem.

Let us assume that the superscript $i\in\{\SRB,\arm\}$ denotes the subsystem index. The local cost function for subsystem $i$ is defined as a quadratic function:
\begin{equation}
    J^{i}\left(\xi^{i}(\cdot)\right):=\mathcal{L}^{i}_{\terminal}\left(x^{i}_{t+N|t}\right) + \sum_{k=0}^{N-1} \mathcal{L}^{i}_{\stage}\left(x^{i}_{t+k|t},u^{i}_{t+k|t}\right),
\end{equation}
where $\mathcal{L}^{i}_{\terminal}(x^{i}_{t+N|t}):=\|x^{i}_{t+N|t} - x^{i,\reff}_{t+N|t}\|_{P^{i}}^{2}$ and $\mathcal{L}^{i}_{\stage}(x^{i}_{t+k|t},u^{i}_{t+k|t}):=\|x^{i}_{t+k|t} - x^{i,\reff}_{t+k|t}\|_{Q^{i}}^{2} + \|u^{i}_{t+k|t}\|_{R^{i}}^{2}$ represent the terminal and stage costs, respectively, for some positive definite matrices $P^{i}$, $Q^{i}$, and $R^{i}$ and some desired state trajectory $x^{i,\reff}(\cdot)$ over the control horizon. Here, we use the standard notation $\|z\|_{Q}^{2}:=z^\top Q\,z$ for any positive definite $Q$ matrix. Similarly, the cost function associated with the interaction wrench is defined as:
\begin{equation}
    J^{\inte}(\lambda^{\inte}(\cdot)) := \sum_{k=0}^{N-1} \mathcal{L}_{\stage}^{\inte}\left(\lambda^{i}_{t+k|t}\right),
\end{equation}
with $\mathcal{L}^{\inte}_{\stage}(\lambda^{\inte}_{t+k|t}):=\|\lambda^{\inte}_{t+k|t}\|_{R^{\inte}}^{2}$ and a positive definite matrix $R^{\inte}$. In Section \ref{sec:Experiments}, we demonstrate that the proposed NMPC formulation is solved in real time at 60 Hz using a receding horizon strategy, where the initial conditions are updated at each time step as $x^{\SRB}_{t|t}=x^{\SRB}(t)$ and $x^{\arm}_{t|t}=x^{\arm}(t)$. The optimal control problem in \eqref{eq:unified_NMPC} is inherently nonlinear, owing to the nonlinear dynamics of both the SRB and the arm subsystems, as well as the holonomic equality constraints that couple them. Numerical and computational details are provided in Section \ref{sec:Setup}. The optimal torque trajectory $u^{\arm}_{t|t}$ is directly applied to the arm for torque-level control, while the predicted SRB state $x^{\SRB}_{t+1|t}$ and optimal GRFs $u^{\SRB}_{t|t}$ are forwarded to a low-level nonlinear controller to compute the joint-level torques required for locomotion (see Fig. \ref{fig:Overview}).


\vspace{-1em}
\subsection{Low-level Nonlinear WBC for Locomotion}
\label{sec:WBC}

At the low level of the control strategy for the locomotion task, we employ a nonlinear WBC to enforce the full-order dynamics of the robot to track the optimal trajectories generated by the high-level NMPC for the SRB subsystem. Specifically, we adopt the nonlinear WBC algorithm developed in  \cite{Randy_Paper_LCSS,pandala2022robust}, formulated as a real-time QP that runs at 500 Hz. This QP solves for the joint-level torques and the GRFs for the full-order model. The WBC formulation ensures that the full-order model GRFs closely follow the optimal GRFs prescribed by the NMPC and integrates a state-tracking mechanism using virtual constraints \cite{Jessy_Book}. 

The real-time QP can be expressed as follows \cite{Randy_Paper_LCSS}:
\begin{alignat}{4}\label{eq:QP}
    &\min_{(u^{\full},f^{\full},\delta)} \,\,\,&& \frac{\gamma_{1}}{2}\|u^{\full}\|^2 + \frac{\gamma_{2}}{2}\|f^{\full} - f^{\des}\|^2 && + \frac{\gamma_{3}}{2}  \|\delta\|^2 \nonumber\\
    &\textrm{s.t.} && \ddot{y} + K_{D}\,\dot{y} + K_{P}\,y = \delta && \textrm{(Output Dynamics)}\nonumber\\
    & && \ddot{r}^{\textrm{st}} = 0 \nonumber && \textrm{(No slippage)}\\
    & && u^{\full}\in\mathcal{U}^{\full},\quad f^{\full}\in\mathcal{FC} && \textrm{(Feasibility)},
\end{alignat}
where $u^{\full}$ and $f^{\full}$ represent the joint-level torques and GRFs for the floating-base full-order locomotion model, $\gamma_{1}$, $\gamma_{2}$, and $\gamma_{3}$ are positive weighting factors, and the desired force profile $f^{\des}(t)$ is provided by the high-level NMPC. The virtual constraints, denoted by $y:=y^{a}-y^{\des}$, encode position tracking, where the controlled variables $y^{a}$ consist of the quadruped's CoM position, Euler angles, and Cartesian positions of the swing feet. The desired trajectory $y^{\des}(t)$  includes the NMPC-prescribed CoM position and Euler angles, and the desired swing foot trajectories are generated using B\'ezier polynomials that interpolate between the current foothold and the next one, computed according to Raibert's heuristics  \cite{raibert1986legged}. The virtual constraints are imposed through the desired output dynamics using positive gains $K_{P}$ and $K_{D}$, with $\delta$ serving as a defect variable to ensure feasibility of the QP. The no-slip condition at the stance feet is enforced by $\ddot{r}^{\textrm{st}}=0$, where $r^{\textrm{st}}$ represents the Cartesian coordinates of the stance feet. The feasibility conditions for joint torques and GRFs are encoded as inequality constraints within the admissible torque set $\mathcal{U}^{\full}$ and the linearized friction cone $\mathcal{FC}$. 


\vspace{-1em}
\section{Experiments}
\label{sec:Experiments}

This section details the design of the proposed NMPC controller and provides a comprehensive evaluation through both physics-engine simulations and hardware experiments.



\begin{figure*}[t]
    \centering
    \includegraphics[width=1.00\linewidth]{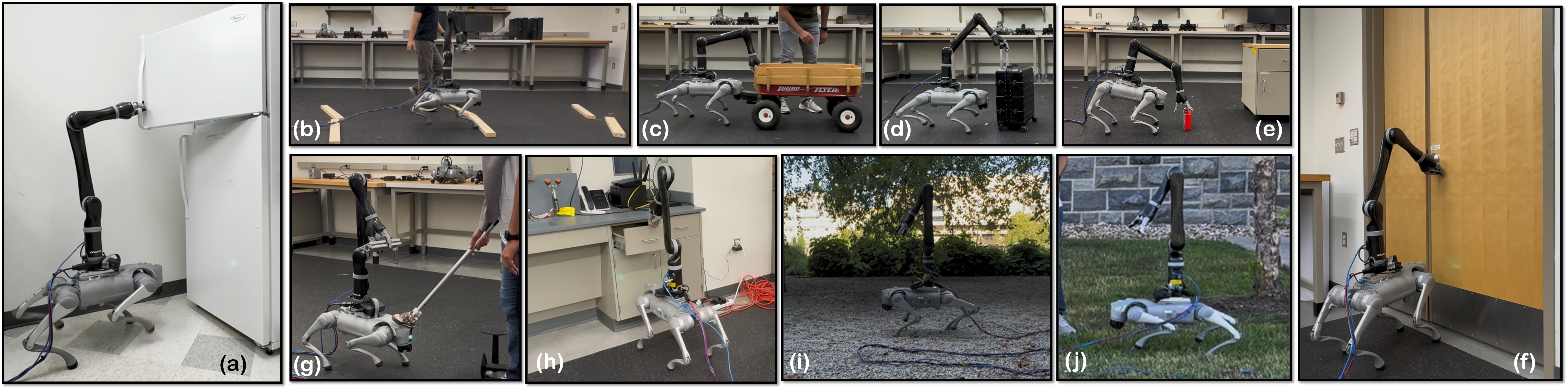}
    \vspace{-2em}
    \caption{Snapshots of experiments demonstrating the deployment of the proposed decomposition-based NMPC algorithm integrated with the low-level nonlinear WBC across diverse loco-manipulation tasks: (a) opening a fridge door and picking up a bottle; (b) walking over wooden blocks while carrying a 1 kg payload; (c) pushing a 13 kg wagon cart loaded with an 11.4 kg payload (125\% of the Go2 robot's mass); (d) pulling a carry-on suitcase; (e) picking up a box from the ground and placing it on a table; (f) opening a lab door without a spring mechanism and passing through it; (g) in-place trotting while carrying a 1 kg payload and subject to external push disturbances; (h) opening a drawer and grasping a penholder cup; and (i-j) walking on gravel and grass surfaces.}
    \vspace{-1.5em}
    \label{fig:Snaphosts}
\end{figure*}


\vspace{-1em}
\subsection{Setup and Controller Synthesis}
\label{sec:Setup}

This work employs the Unitree Go2 quadrupedal robot and the Kinova Gen2 4-DoF manipulator as the hardware platform for both numerical and experimental validation of the proposed loco-manipulation NMPC framework. The Go2 is a 15.0 kg quadruped with a standing height of 0.28 m and 18 DoFs, including 12 actuated joints---three per leg (hip pitch, hip roll, and knee pitch). The Kinova Gen2 is a 4.4 kg manipulator featuring four torque-controlled DoFs and a 3-fingered gripper with position and velocity control. A Sony DualShock 4 controller, operated by a human user, is used to provide real-time reference commands to the NMPC, including the desired base velocity, orientation, and arm configuration. The NMPC algorithm was executed offboard in a multithreaded fashion on a desktop PC equipped with a 12th Gen Intel® Core™ i9-12900F CPU and 64 GB of RAM. Communication between the PC and the robots was handled via two Ethernet switches operating in a LAN configuration. For numerical simulations, the RaiSim physics engine \cite{RAISIM} was used to validate the performance of the proposed control framework. 

\textbf{NMPC Hyperparameters:} The hyperparameters of the NMPC for the SRB component are chosen as $Q^{\SRB}=\textrm{block diag}\{Q^{\SRB}_{p},Q^{\SRB}_{\dot{p}},Q^{\SRB}_{\theta},Q^{\SRB}_{\omega}\}$ with  $Q^{\SRB}_{p}=\diag\{1e8,1e8,8e8\}$, $Q^{\SRB}_{\dot{p}}=\diag\{1e6,1e6,1e6\}$, $Q^{\SRB}_{\theta}=\diag\{1e8,1e8,5e8\}$, and $Q^{\SRB}_{\omega}=\diag\{5e4,5e4,5e4\}$. The terminal cost and control penalty are set to $P^{\SRB}=20\,Q^{\SRB}$ and $R^{\SRB}= 100\,\identity_{3\times3}$, respectively. For the arm subsystem, the weighting matrix is defined as $Q^{\arm}=\textrm{block diag}\{Q^{\arm}_{p^{b}}, Q^{\arm}_{\theta^{b}}, Q^{\arm}_{q^s}, Q^{\arm}_{\dot{p}^{b}}, Q^{\arm}_{\dot{\theta}^{b}}, Q^{\arm}_{\dot{q}^s}\}$ where the nonzero components are specified as $Q^{\arm}_{q^s} = \diag\{8.5e7, 8e7, 8e6, 3e5\}$ and $Q^{\arm}_{\dot{q}_s} = \textrm{diag}\{8e5, 8e5, 8e4, 8e5\}$, and all base-related terms are set to zero, i.e., $Q^{\arm}_{p^{b}}=Q^{\arm}_{\theta^{b}}=Q^{\arm}_{\dot{p}^{b}}=Q^{\arm}_{\dot{\theta}^{b}}=0_{3\times3}$ since they are already regulated by the SRB subsystem. The terminal and control weights for the arm are set to $P^{\arm}=10\,Q^{\arm}$ and $R^{\arm}= 100\,\identity_{4\times4}$. Finally, the weighting matrices for interaction forces and torques are selected as $R^{\inte} =  \textrm{block diag}\{R^{\inte}_f,R^{\inte}_{\tau}\}$ with $R^{\inte}_f = 500\, \identity_{3\times3}$ and $R^{\inte}_{\tau} = 5\,\identity_{3\times3}$. 

\textbf{Real-Time Computation:} The control horizon was set to $N = 8$ with a sampling time of $T_{s} = 16.67$ ms (corresponding to a control frequency of 60 Hz). We used the trot gait for the quadrupedal robot with a step time of 200 ms. The friction coefficient $\mu$ was varied between 0.4 and 0.8 to account for different terrain conditions shown in Fig. \ref{fig:Snaphosts}. The nonlinear optimization problem in (\ref{eq:unified_NMPC}) involves 464 decision variables, including $\xi^{\SRB}(\cdot)$, $\xi^{\arm}(\cdot)$, and $\lambda^{\inte}(\cdot)$, as defined in (\ref{eq:unified_NMPC}), along with the initial states of the SRB and arm, which are constrained to match the measured state feedback at time $t$. The NLP was implemented using the CasADi framework \cite{CasADI} with the IPOPT solver \cite{IPOPT}. Each NMPC iteration was limited to 10 solver iterations, using the previous solution as the warm-start initial guess. The real-time loco-manipulation NMPC formulation in (\ref{eq:unified_NMPC}) is inherently nonconvex and is solved online without any approximation to the nonlinear SRB dynamics. To reduce computational complexity, the manipulator's Coriolis term is neglected, and its inertia matrix is assumed constant over the control horizon, updated at each NMPC cycle to reflect the latest system configuration. Gravity and holonomic constraint terms remain nonlinear. Table \ref{tab:solvetime} summarizes the NMPC computation time distribution across time intervals (in milliseconds) for two challenging scenarios involving rough terrain and external disturbances. Furthermore, no infeasibility was observed during the experiments. The low-level WBC in \eqref{eq:QP} was solved using the qpSWIFT solver \cite{qpSWIFT} at 500 Hz with the following hyperparameters: $\gamma_{1} = 1$, $\gamma_{2} = 1e7$, $\gamma_{3} = 1e6$, $K_P = 400$, and $K_D = 40$. 

We observed that without the proposed decomposition strategy in \eqref{eq:unified_NMPC}, integrating full SRB and arm dynamics into a unified model renders real-time NMPC computation with CasADi impractical, necessitating major simplifications such as linearizing mass-inertia and Coriolis terms. In contrast, the proposed method handles SRB nonlinearities while simplifying only parts of the arm dynamics, preserving the locomotion template's effectiveness through subsystem-specific cost function decomposition, as discussed in Section \ref{sec:Contributions}.

\begin{table}[t]
\centering
\caption{Distribution of NMPC solve times versus intervals (ms)}
\label{tab:solvetime}
\vspace{-1.0em}
\begin{tabular}{|c|c|c|c|c|c|}
\hline
Experiment & $12-13$ & $13-14$ & $14-15$ & $15-16$ & $\geq16 $\\
\hline
Rough terrain & $4.3\%$ & $87.7\%$ & $6.3\%$ & $0.7\%$ & $1\%$\\
External push & $4.9\%$ & $85.6\%$ & $8.4\%$ & $0.4\%$ & $0.7\%$\\
\hline
\end{tabular}
\vspace{-1.7em}
\end{table}


\vspace{-1em}
\subsection{Hardware Experiments}
\label{sec:Hardware_Experiments}

In this section, we classify and summarize the experiments conducted throughout the study. All experimental videos are available online \cite{Ruturaj_RAL_YouTube}. 

\textbf{Robust loco-manipulation on various terrains:} In the first phase of experiments, the robot is commanded to walk over diverse terrains---including wooden blocks, grass, and gravel---as shown in Fig.~\ref{fig:Snaphosts}(b, i, and j). The objective is to evaluate the robustness of the composite robot's gait while a human operator supplies a time-varying reference velocity trajectory for the robot's base. During these experiments, the NMPC receives a fixed arm configuration as input. The rough terrain scenario is replicated in the lab using 5 cm-tall wooden blocks (17.9\% of the Go2's standing height), which are unmodeled in the NMPC---i.e., the robot has no prior knowledge of their presence and does not rely on any perception system to detect them. Despite this, the NMPC successfully stabilizes both the SRB and arm states, demonstrating robust performance under unstructured terrain conditions.

\begin{figure}[t]
    \centering
    \includegraphics[width=0.9\linewidth]{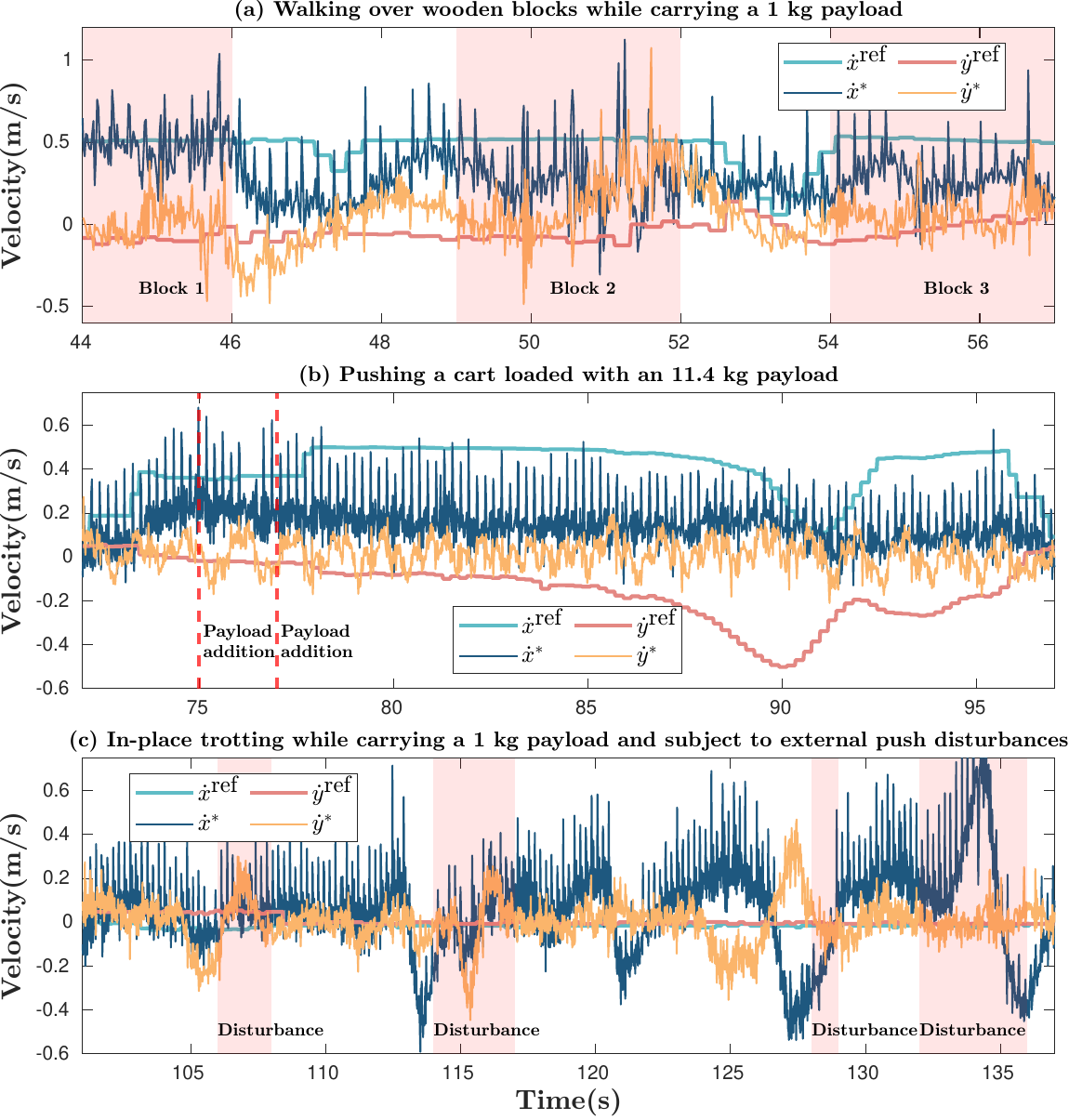}
    \vspace{-1.3em}
    \caption{Plot of the reference base velocity trajectories $(\dot{x}^{\reff},\dot{y}^{\reff})$ commanded via joystick, alongside the optimal velocity trajectories $(\dot{x}^{\star},\dot{y}^{\star})$ prescribed by the high-level NMPC for: (a) walking over wooden blocks while carrying a 1 kg payload (see Fig. \ref{fig:Snaphosts}(b)); (b) pushing a 13 kg wagon cart with an 11.4 kg payload (125\% of the Go2's mass) (see Fig. \ref{fig:Snaphosts}(c)); and (c) in-place trotting while carrying a 1 kg payload under external push disturbances (see Fig. \ref{fig:Snaphosts}(g)). In (a), the shaded region indicates intervals where the robot is walking over three wooden blocks. In (b), the dashed lines denote time instances when the payload is incrementally added. In (c), the shaded regions represent periods when external push disturbances are applied.} 
    \vspace{-1.7em}
    \label{fig:SRB_plots}
\end{figure}

\textbf{Robust loco-manipulation with unknown payloads and subject to external disturbances:} In the second phase of experiments, the robot repeats the rough terrain locomotion task while carrying an \textit{unknown} 1 kg payload using the manipulator (see Fig. \ref{fig:Snaphosts}(b)). Additionally, we assess the robustness of the loco-manipulation system by subjecting the robot---while carrying the same 1 kg payload---to external push disturbances applied by a human bystander (see Fig. \ref{fig:Snaphosts}(g)). The commanded base speed trajectories, along with the optimal trajectories generated by the NMPC, for these two experiments are shown in Fig. \ref{fig:SRB_plots}(a) and (c). The results demonstrate that the NMPC can robustly stabilize the system despite unknown payloads, external disturbances, and unstructured terrain.


\textbf{Pushing and pulling objects with unknown payloads:} In the third phase of experiments, the robot is commanded to push a 13 kg wagon cart carrying an unknown payload of 11.4 kg, as shown in Fig.~\ref{fig:Snaphosts}(c). The payload is added incrementally by a human bystander during the task. It is worth noting that the total mass of the composite robot is approximately 19.4 kg, and in this task, it pushes an \textit{unmodeled} object with a mass equivalent to about 125\% of its own weight, highlighting the system’s robustness to significant external uncertainties. The optimal state trajectories prescribed by the NMPC are illustrated in Fig. \ref{fig:SRB_plots}(b). The robot is also capable of pulling the same wagon cart. In addition, we demonstrate a practical application in which the robot pulls a 3.7 kg carry-on suitcase, as shown in Fig. \ref{fig:Snaphosts}(d). 


\begin{figure}[t]
    \centering
    \includegraphics[width=0.85\linewidth]{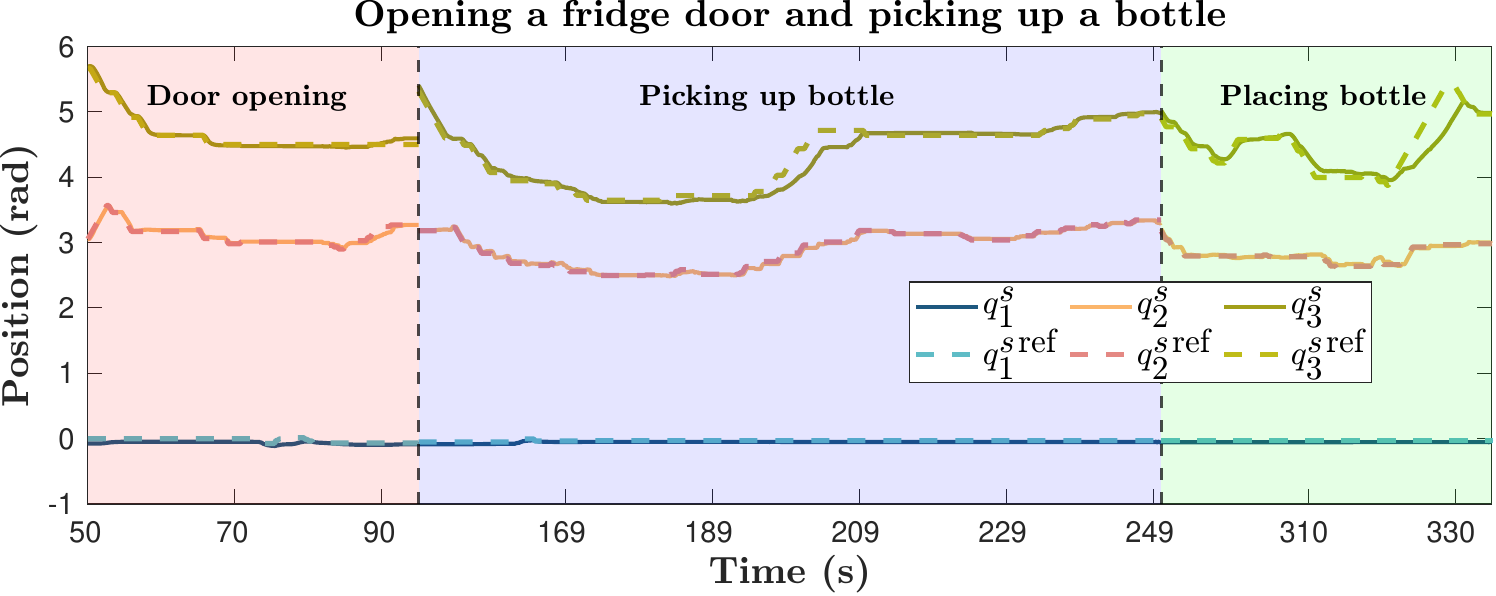}
    \vspace{-1.3em}
    \caption{Plot of the reference arm position trajectories commanded by the user, alongside the actual arm position trajectories for three out of four joints of the Kinova arm (excluding wrist motion), during (a) fridge door opening, (b) bottle grasping, and (c) bottle placement on an adjacent countertop.} 
    \vspace{-1.7em}
    \label{fig:arm_plots}
\end{figure}


\textbf{Pick-and-place tasks:} In the fourth set of experiments, the robot performs object manipulation tasks that involve picking up items from one location and placing them in another. These tasks are designed to evaluate the effectiveness of the proposed NMPC framework in coordinating whole-body locomotion and arm motion under diverse loco-manipulation scenarios. Specific examples include retrieving a puzzle cube or football from a table and placing them into a box on the ground, as well as lifting a first-aid kit from the ground and placing it on a table (see Fig. \ref{fig:Snaphosts}(e)). In these experiments, the user provides base pitch and yaw commands via a joystick, while the NMPC ensures the stability of the quadrupedal robot and generates torque commands to track the desired joint-space trajectories for the arm subsystem, also specified through the joystick.

\textbf{Opening fridge, door, and drawer:} The final set of experiments demonstrates common teleoperated loco-manipulation tasks representative of home environments, highlighting the practical applicability of the proposed NMPC framework. The first task consists of opening a fridge door, grasping a bottle, placing it on an adjacent countertop, and closing the fridge door, as shown in Fig. \ref{fig:Snaphosts}(a). A key challenge in this task is that the fridge door requires substantial force to open, which causes noticeable displacement of the quadrupedal robot. To counteract this, the robot uses the fridge surface as frontal support to apply the required force. In the second task---door opening (see Fig. \ref{fig:Snaphosts}(f))---the robot rotates the handle of a locked lab door, pushes it open (in the absence of a spring or damping mechanism), and then passes through the doorway. The third task involves opening a drawer, grasping a penholder cup, closing the drawer, and placing the cup on a nearby tabletop (see Fig. \ref{fig:Snaphosts}(h)). Throughout these experiments, the user provides velocity-level joystick commands to the quadrupedal base, individual arm joints, and the gripper. The NMPC ensures stability and enables teleoperated manipulation involving end-effector force application and interaction with unmodeled objects. Figure \ref{fig:arm_plots} shows the reference position trajectories provided by the user alongside the actual position trajectories of the arm subsystem during the fridge experiment.


\vspace{-1em}
\subsection{Comparison and Quantitative Analysis}
\label{sec:Comparison}

We conduct extensive simulations to numerically evaluate the efficacy and robustness of the proposed decomposition-based NMPC algorithms in comparison with a commonly used baseline MPC approach for loco-manipulation. As highlighted in Section~\ref{sec:Related_work}, a unified NMPC framework---such as the one presented in \cite{Sleiman_RAL, Sleiman_Science}---considers the quadrupedal robot's centroidal dynamics along with the full-body kinematics of the arm subsystem. Our evaluation focuses on the robust locomotion performance of the Go2 quadrupedal robot with the Kinova arm at a target speed of 0.5 m/s across 200 randomly generated terrains, each populated with 20 blocks varying in height (1-3 cm), width (10-20 cm), and length (1-7 m).


\begin{figure}[t]
    \centering
    \includegraphics[width=0.85\linewidth]{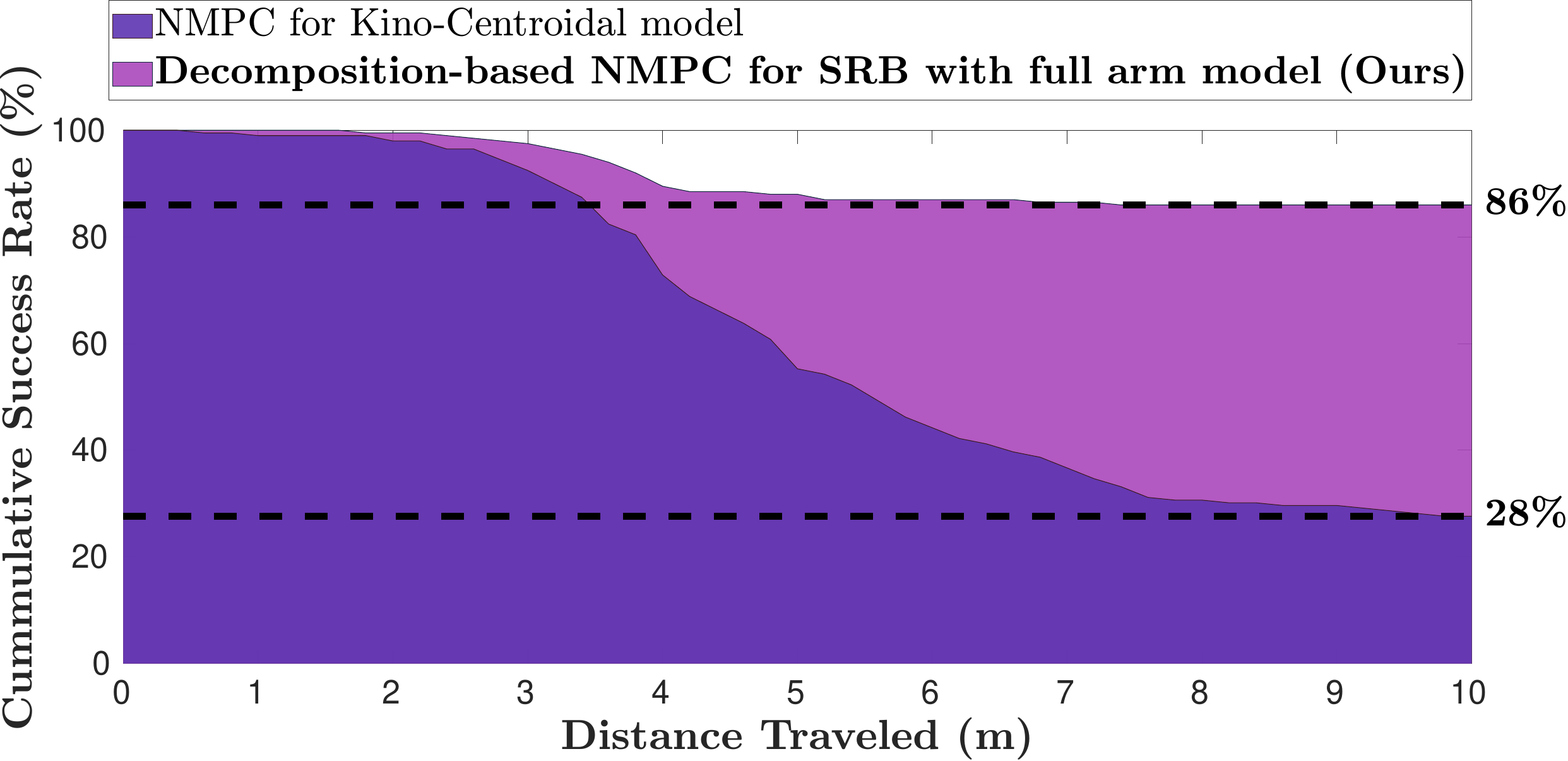}
    \vspace{-1.3em}
    \caption{Success rate comparison between the proposed decomposition-based NMPC, integrating SRB dynamics with the full-order arm model, and the baseline kino-centroidal NMPC approach as a function of traveled distance.} 
    \vspace{-1.4em}
    \label{fig:Comparison_plot}
\end{figure}


For comparison, we adopt the layered control structure shown in Fig. \ref{fig:Overview}, where the proposed decomposition-based NMPC approach is replaced with the kino-centroidal NMPC framework from \cite{Sleiman_RAL}. In the kino-centroidal NMPC technique, optimal trajectories are generated for centroidal momentum, base position and orientation, and arm joint positions. The optimal base position and orientation trajectories are then passed to our low-level nonlinear WBC for tracking with the full-order model, consistent with the proposed method. However, the arm’s optimal joint-space trajectories are tracked using a proportional-derivative (PD) controller to generate torque commands. Notably, unlike our approach, the kino-centroidal NMPC does not compute optimal torque trajectories for the arm subsystem. In this comparison, both NMPC algorithms are executed at 60 Hz, while the low-level WBC operates at 500 Hz. Figure \ref{fig:Comparison_plot} illustrates the success rate of each approach for the composite robot as a function of traveled distance. Here, success is defined as the robot reaching 10 m while satisfying stability bounds. From Fig. \ref{fig:Comparison_plot}, the overall success rates for the proposed decomposition-based NMPC and the kino-centroidal NMPC are 86\% and 28\%, respectively. These results indicate that the proposed decomposition-based NMPC significantly outperforms the kino-centroidal method, particularly in integrating small quadrupedal robots such as Go2 with heavier manipulators like the Kinova arm.


\vspace{-1em}
\section{Conclusions}
\label{sec:Conclusions}

This paper presented a computationally efficient NMPC framework for loco-manipulation tasks involving quadrupedal robots augmented with manipulators whose dynamics are non-negligible relative to the base template model. The proposed approach employs a decomposition strategy that couples locomotion template models with full-order arm dynamics for torque control. This decomposition enables a unified and efficient real-time NMPC formulation that simultaneously optimizes the reduced-order base and full-order arm trajectories. Reduced-order locomotion trajectories are tracked using a low-level nonlinear WBC, while NMPC-generated torque commands are applied directly to the arm. The effectiveness of the proposed approach was validated through extensive numerical simulations and hardware experiments involving diverse loco-manipulation tasks performed by the Go2 quadrupedal robot augmented with a 4-DoF Kinova arm, under conditions of payload uncertainty, external disturbances, and rough terrain. Unlike prior works that rely on small quadrupedal robots equipped with lightweight, position-controlled arms, this work introduces a model-based approach that accounts for the full-order dynamics of a heavy, torque-controllable Kinova arm mounted on the compact Go2 robot, with an arm-to-base mass ratio of 29\%. Compared to kino-centroidal NMPC baselines designed for heavier quadrupeds with lower arm-to-base mass ratios, the proposed approach achieved a 58\% higher robust locomotion success rate.

This study validated the approach using the Go2 robot with a 4-DoF Kinova arm, selected due to mass constraints and its torque control capabilities. The limited DoF increases operator effort in complex tasks. Future work will explore higher-DoF arms to improve manipulability.





\vspace{-1em}
\bibliographystyle{IEEEtran}
\bibliography{references}

\end{document}